\documentclass[sigconf]{acmart}

\usepackage{icaif-2022-scoring}

\AtBeginDocument{%
  }

\setcopyright{acmcopyright}
\copyrightyear{2022}
\acmYear{2022}
\acmDOI{XXXXXXX.XXXXXXX}

\acmConference[XAI-FIN-2022]{Workshop on Explainable AI in Finance}{November 2, 2022}{New York, NY}

\begin{document}

\title{Evaluation Metrics for Symbolic Knowledge Extracted from Machine Learning Black Boxes: A Discussion Paper}

\author{Federico Sabbatini}
\authornote{Corresponding author.}
\email{f.sabbatini1@campus.uniurb.it}
\orcid{0000-0002-0532-6777}
\affiliation{%
	\institution{University of Urbino}
	\city{Urbino}
	\country{Italy}
}

\author{Roberta Calegari}
\email{roberta.calegari@unibo.it}
\orcid{0000-0003-3794-2942}
\affiliation{%
  \institution{\textsc{Alma Mater Studiorum}---University of Bologna}
  \city{Bologna}
  \country{Italy}
}

\renewcommand{\shortauthors}{F. Sabbatini, R. Calegari}

\begin{abstract}
  As opaque decision systems are being increasingly adopted in almost any application field, issues about their lack of transparency and human readability are a concrete concern for end-users.
  Amongst existing proposals to associate human-interpretable knowledge with accurate predictions provided by opaque models, there are rule extraction techniques, capable of extracting symbolic knowledge out of an opaque model.
  However, how to assess the level of readability of the extracted knowledge quantitatively is still an open issue.
  Finding such a metric would be the key, for instance, to enable automatic comparison between a set of different knowledge representations, paving the way for the development of parameter autotuning algorithms for knowledge extractors.
  In this paper we discuss the need for such a metric as well as the criticalities of readability assessment and evaluation, taking into account the most common knowledge representations while highlighting the most puzzling issues.
\end{abstract}



\maketitle

\section{Introduction}

Artificial Intelligence (AI) for finance is a reality including, but not limited to, consumer credit, credit risk, and anti-money laundering applications \cite{aziz2019machine}. 
Usually, to obtain high predictive capabilities, machine learning techniques exploited for predictions rely on sub-symbolic techniques, that offer advantages related to performances but also come along with some drawbacks~\cite{rocha2012far}.
The main drawback lies in the opacity of these models, which are obscure in the decisions they make and lack human interpretability.
For such a reason models of this kind are also called \emph{black boxes} (BB).

When dealing with critical applications -- such as the finance ones -- lack of human awareness is even more unacceptable and makes BB models impracticable. 
Finance is a domain where trustworthiness must be ensured \cite{ng2015ethical}. Unique ethical, legal, and regulatory challenges arise as decisions can immediately impact people's lives \cite{weiss2021business}. 
However, it is still unclear how to implement and test trustworthy AI systems in practice. Following the definition of the High-Level Expert Group on AI \cite{egtai}, trustworthy AI should satisfy three main pillars: comply with all applicable laws and regulations (\emph{lawfulness}), adhere to ethical principles and values (\emph{ethicality}), and be safe, secure and reliable (\emph{robustness}). However, how to translate these principles into practice is still an open question \cite{mittelstadt2019principles}.
A possible step towards trustworthy AI is to develop explainable AI (XAI), as explainability is one of the 7 more specific requirements that refine the main pillars \cite{egtai}. XAI aims to create insight into how and why AI models produce predictions while maintaining high predictive performance levels.
%

%
%
%
%
%
%

Different strategies are present in the literature to tackle XAI via explainable models~\cite{guidotti2018survey}:
\begin{inlinelist}
	\item human-interpretable models, such as shallow decision trees~\cite{Rudin2019}; or
	\item symbolic knowledge-extraction (SKE) techniques to the BB~\cite{KENNY2021103459}.
\end{inlinelist}
While \textit{i)} is not always applicable, \textit{ii)} is gaining a momentum. The amount of SKE algorithms available in the literature is constantly increasing, also thanks to the software support provided by object-oriented frameworks for machine learning as well as to the development of dedicated frameworks (e.g., \psyke{}~\cite{psyke-woa2021,psyke-ia2022,psyke-extraamas2022}).
%
%
Amongst the application areas of SKE there are financial~\cite{baesens2001building,baesens2003using,steiner2006using} and medical fields~\cite{bologna1997three,franco2007early,hayashi2000comparison} and others~\cite{setiono2011rule,sabbatini22LPFSKE,azcarraga2012keyword,hofmann2003rule}.
Examples of extractors generally achieving good results are \cart{}~\cite{breiman1984classification}, \trepan{}~\cite{craven1996extracting}, \gridex{}~\cite{gridex-extraamas2021} and \gridrex{}~\cite{gridrex-kr2022}.
Of course a number of alternatives are present in the literature~\cite{craven1994using,huysmans2006iter,barakat2005eclectic,martens2007comprehensible}, depending on the task at hand.

A common trait of every SKE extractor is the output, which has to be some sort of human-interpretable knowledge.
The knowledge usually follows a list or tree representation, where each item (leaf) of the list (tree) corresponds to a specific rule having predictive power.
However, how to select the best SKE extractor for the case at hand -- possibly as an automatic task -- is still an open issue and the level of human readability of the extracted knowledge should be evaluated via a fair metric \cite{Sovrano2022,Mulder2021}.
The level of human-readability of the extracted knowledge is bounded to the number of rules it contains and, recursively, to the readability of each rule, in turn.
Indeed, rules may follow several formats with different levels of conciseness even when having the same semantics.
For this reason, the quantitative assessment of knowledge readability -- through an unbiased score and relative metric -- is a quite challenging task.
The lacking of such a metric prevents the design and implementation of procedures to automatically tune extractor parameters since an unbiased comparison of extractors' outputs is not possible.
Accordingly, in this paper we provide a detailed discussion about the readability assessment, highlighting the most challenging issues. The discussion is clustered based on the knowledge representation of the extraction algorithm.

\section{Evaluating Extracted Knowledge}

Following research discussions on the topic~\cite{garcez2001symbolic,tran2013knowledge}, the most common indicators to evaluate the knowledge extracted via SKE techniques are:
\begin{inlinelist}
	\item the predictive performance of the knowledge rules w.r.t.\ the data or the BB predictions (the latter is called \emph{fidelity});
	\item the readability level from a human perspective; and
	\item the input space coverage (\emph{completeness}).
\end{inlinelist}
Having good scores according to these 3 dimensions ensures that the extracted knowledge represents well the underlying domain, that it is interpretable for humans, and that it may be used as a surrogate model to draw predictions. 

It is quite easy to evaluate the knowledge predictive performance with the same scoring function adopted to evaluate the underlying BB.
Adequate scoring metrics are, for instance, the accuracy and F$_1$ scores for classification tasks and the mean absolute/squared error and R$^2$ score for regression tasks.
This indicator is pretty relevant since it represents the best way to assess if the knowledge is suitable to draw predictions in lieu of the BB.

The input space coverage can also be measured trivially, e.g., by observing how many data set instances are covered by the extracted rules, or by performing some kind of equal-spaced or random sampling inside the input space and then checking how many data points are covered.
The latter solution is more demanding for high-dimensional data sets and may have little significance in the case of sparse data sets (characterised by empty regions).
This indicator is also important because extremely human readable knowledge (e.g., composed of a single, simple rule) having excellent predictive performance (e.g., 100\% accuracy) is useless if it covers an infinitesimally small region of the input feature space.

Human-readability assessment and evaluation could turn out not to be so trivial.
First of all, we have to distinguish between two different levels of readability, both related to the presentation of the extracted knowledge to users.
On the one hand, we can consider the \emph{macrolevel} of readability of the knowledge as a whole, related to the knowledge shape and dimension -- e.g., if it is represented as a tree, a list, a table, etc. -- and to the number of leaves, rules, rows/columns present (aka complexity).
On the other hand, we can focus on the \emph{microlevel} of readability, to recursively take into account the same indicators -- shape and complexity -- for each element in the knowledge.
As an example of the microlevel we can consider for instance the kind of preconditions of each element (e.g., oblique rules, \mofn{} tree nodes) and how many variables, constants, and predicates appear inside them.
As can be noticed, this is far from being a trivial task, since equivalent concepts may be represented according to different notations, each one having its conciseness and expressiveness.
In the following, both macrolevel and microlevel are examined in detail, showing that the most challenging issues are related to the latter.

\subsection{Macrolevel: knowledge as a whole}\label{ssec:macro}

As widely acknowledged~(e.g.,~\cite{guidotti2018survey}), the most common knowledge representations, when such a knowledge has to be presented to human users, are decision lists~\cite{freitas2014comprehensible,huysmans2011empirical}, trees~\cite{quinlan1993c4,breiman1984classification} and tables~\cite{sethi2012kdruleex}.
The reason behind their massive exploitation is related to their high level of understandability for humans.

\subsubsection{Knowledge shapes}

\paragraph{Decision lists}

Knowledge having this shape is represented as a list of rules, where each rule has a precondition and a postcondition.
Preconditions usually are conjunctions -- or less frequently disjunctions -- of constraints on the input variables.
Postconditions, on the other hand, are the associated outputs.
Thus, a rule provides output predictions for all and only those input instances satisfying the rule precondition.
The complexity of the list may be easily assessed through the number of contained rules.

Lists may be exhaustive if it is always possible to draw predictions for input instances (i.e., at least one rule can be applied to each input instance), or not.
This characteristic does not affect the knowledge readability, but only the input space coverage.

Furthermore, rules may overlap -- if more than one rule may be applied simultaneously to a single input instance -- or not---if at most one rule can be used.
This feature may affect the overall readability, however, in practice, it can be neglected since knowledge extracted via SKE techniques usually relies on non-overlapping rules, or the overlapping rules have some degree of confidence thanks to which it is possible to select a single rule for each input sample.

Rule lists may be hierarchical if some rules have as a postcondition a sublist of rules.
They can be trivially converted into a non-hierarchical list, so there is no need to consider this specific scenario.

Finally, rule lists may be ordered -- when rule number $n$ is evaluated only if preconditions of rules $1 \dots n-1$ are not satisfied -- or unordered---if, otherwise, any rule can be chosen without taking into account the preceding in the list.
Human readability is strongly impacted by the list ordering since human users must be conscious that a rule can be selected only if all the preceding have been already tested.
In other words, the $n$-th rule implicitly carries along information about the preconditions of other distinct $n-1$ rules, i.e., implicit information that is inherently coded within the ordering.
To give an example, the fourth rule in an unordered set may be something like ``output is $post_4$ if $pre_4$'', and humans can read and apply the rule as is.
We remark that $pre_i$ and $post_i$ are the precondition and postcondition, respectively, associated with the $i$-th rule of the list.
The same rule in an ordered set should be considered by humans as ``output is $post_4$ if $pre_4, not(pre_3), not(pre_2), not(pre_1)$'', where $not(pre)$ is true if $pre$ is not satisfied and the comma operator represents logical conjunction.
Of course, the readability of ordered sets is hindered by this consideration, especially for a large amount of rules.

\paragraph{Decision trees}

Knowledge adhering to this representation differs from rule lists since in this case rules are represented as complete paths from the tree root to the leaves.
Each distinct path from the root to a different leaf is a rule, so the complexity of the knowledge is equal to the number of leaves (that in turn is equal to the number of rules).
Decision trees are usually binary trees storing a constraint on one or more input variables in each internal node and a decision in each leaf.
As a consequence, the precondition of a rule is given by the logical conjunction of the constraints corresponding to the internal nodes.
Conversely, the postcondition is associated with the leaf.

Decision trees are always exhaustive by design since it is always possible to draw predictions for the given instances.
The provided rules are also non-overlapping, because only one leaf may be reached when examining an input sample.

We believe that the best option to analyse decision tree human readability is to convert the tree into a (possibly ordered) rule list, and then stick to the same scoring metrics adopted for lists.

\paragraph{Decision tables}

Knowledge following this representation is provided in tabular form.
Usually, each column (or row) represents a variable and each row (or column) is associated with a rule.
In the following, we associate columns with variables and rows with rules.
The cell obtained by intersecting column $j$ with row $k$ represents the constraint on input variable $j$ in the $k$-th rule.
If variable $j$ is the output variable, then the cell contains the output decision of the rule.
Usually, the output column is the last.

By assuming this notation, it is straightforward to obtain rules having as precondition the conjunction of all the constraints on the same row and as postcondition the contents of the last row cell.
It is as well trivial to obtain the complexity of the table in terms of number of rows (equal to the rule amount).

To enable a fair comparison between decision tables and other kinds of knowledge shapes, the former may be easily converted into (generally unordered) rule lists.

\subsubsection{Macrolevel observations}

\Cref{ssec:macro} may be summarised in a few considerations: decision lists, trees, and tables are easily comparable in terms of human readability, since even if they provide knowledge according to different formats, it is always possible to obtain an equivalent unordered list without predictive losses.
The equivalent list will have different individual rules -- e.g., more complex than those in the original knowledge -- but the knowledge macrolevel complexity will not be affected---e.g., the rule amount has to remain the same.
In this way, it is possible to perform a fair comparison amongst different knowledge shapes by only focusing on the macrolevel complexity and further considerations about the microlevel.
During a macrolevel readability assessment, knowledge having smaller complexity (i.e., fewer rules) should be preferred over other solutions.

\subsection{Microlevel: individual rules}

When extracting knowledge, SKE techniques output rules adhering to different representations for both preconditions and postconditions.
Amongst the most widespread kinds of rule preconditions we can find propositional, \mofn{}~\cite{TowellS91,murphy1991id2}, fuzzy~\cite{HorikawaFU92,Berenji91} and oblique~\cite{Setiono00,SetionoL97} literals.
In any case, preconditions are expressed as a conjunction, or disjunction, of literals having one of these different formats.
On the other hand, postconditions are constant values or functions of the input variables.

\subsubsection{Rule preconditions}

\paragraph{Propositional rules}

The most adopted rule format is the propositional one, where literals are simple predicates expressing equalities (in positive or negative form) or inequalities between the value of an input feature and one or more constants.
The basic notations for equalities and inequalities are $X = u$, $Y \neq v$; $Z \lessgtr w$, where $X,Y,Z$ are input features and $u,v,w$ are constant values.
Basic literals may be extended to interval and set inclusion/exclusion, to group together disjunctions or conjunctions of a set of literals.

This first step to achieve better conciseness comes along with a major readability issue.
Indeed, a knowledge may contain two rules having the same output and only one precondition each, composed of a basic literal involving the same variable but compared to different constants, for instance ``output is $A$ if $color=blue$'' and ``output is $A$ if $color=red$''.
Rules of this kind may be collapsed into a single rule: ``output is $A$ if $color=blue$ or $color=red$''.
But this is not the only solution, since the rule ``output is $A$ if $color \in \{blue,red\}$'' is also equivalent.

Starting from a trivial example it is easy to notice that even one categorical feature assuming at least 2 distinct values leads to 3 different knowledge representations, all equivalent.
The readability of an output knowledge should be proportional to the effort required by a human user to understand and apply its rules.
In this case, the first knowledge has 2 rules, 2 variable occurrences and as many predicates and constants.
The second has 1 rule and the same values for other indicators.
The third knowledge has 1 rule, 1 variable occurrence, 1 predicate, and 2 constant values.
Obviously, it is preferable to have 1 rule, and this is decided via macrolevel considerations.
Amongst the second and third knowledge rules, there is not a substantial readability difference, since both of them imply the human user to evaluate the $X$ variable and compare it with two constant values.
As a result, we obtain that a raw counting of variable occurrences, predicates, and constants is not suitable to express microlevel readability, even though they are perfect indicators for rule conciseness (in this case the fewer, the better).
Furthermore, one may observe that the more a rule adopts concise literals, the more it is difficult to understand the rule and to associate a readability score to it, as shown in the following.

\paragraph{\mofn{} rules}

Literals may be expressed as groups, where the group is satisfied if at least one of the possible constraints is satisfied, as shown in the previous paragraph.
This definition may be further relaxed, and a group of $N$ literals can be considered as satisfied if at least $M$ constraints are satisfied ($M\leq N$).
\mofn{} rules are exactly defined in this manner.
They enable a much higher degree of conciseness, however, our opinion is that human users encounter more difficulties in understanding the rule.
As an example, one may think about a 4-of-9 rule.

Large $N$ values also obfuscate the real impact of individual variables on the final prediction.
Furthermore, it is quite challenging mapping \mofn{} rules to readability scores, since their readability decreases w.r.t.\ both $M$ and $N$.
For instance, 1-of-5 rules are more easily understandable than 3-of-5 or 1-of-10 rules.
But analogous claims about 5-of-9 vs. 4-of-10 rules cannot be easily stated.

\paragraph{Fuzzy rules}

If sets and intervals in preconditions are substituted with string labels and these labels correspond to the belonging to a defined set/interval within a certain confidence degree, rules are called fuzzy.
For instance, ``$X \in [0.3, 0.6]$'' could be equivalent to ``$X$ is $medium$'', by assuming the $X$ variable ranging in the [0, 1] interval.
The conciseness of this notation allows literals to have more flexible intervals (e.g., also [0.45, 0.65] could be considered as $medium$, with a different degree of confidence).
Outputs associated with fuzzy rules also have a confidence degree, depending on the one of corresponding preconditions.
As a consequence, human users reading a fuzzy rule need to
\begin{inlinelist}
	\item be aware of the underlying fuzzy labelling semantics;
	\item assign a degree of confidence to each literal in the rule precondition;
	\item read the postcondition and calculate the corresponding confidence.
\end{inlinelist}
We believe that it would be far less readable than a direct, simple basic notation.
Since it is not possible to numerically assess the introduced human-interpretability hindrance, it is also hard to assign a readability score to fuzzy rules.

\paragraph{Oblique rules}

When dealing with numerical attributes, rule literals may be expressed as inequalities involving a linear combination of the input variables and a constant threshold.
On the input feature space, these rules represent oblique hyperplanes.

Knowledge conciseness benefits from the application of oblique rules since composite features are enabled, however, in our opinion, human readability pays a high price.
One may think about the description of a concrete object to be classified.
Such an object may be represented through weight, volume, and position.
Classification involving input features separately is easy for humans.
The same is not true if human users have to consider, for instance, a linear combination of weight and position.
For this reason, the same final considerations about the fuzzy rule readability hindrance also hold for oblique rules.

\subsubsection{Rule postconditions}

\paragraph{Constant values}

The most simple and human-readable postcondition is represented by a constant value, suitable for both classification and regression tasks, since it may be a string label as well as a numerical value.
However, constant values are not always the best solutions when dealing with regression data sets, especially if the output feature is strongly correlated to the input variables.
The reason is that constant values introduce an undesired discretisation that usually leads to the creation of many rules with poor predictive performance.

\paragraph{Non-constant postconditions}

Also for postconditions, as for preconditions, there is the possibility to choose more expressive representations, at the expense of human readability.
Indeed, constant predictions can be substituted with linear combinations or other kinds of functions involving (a subset of) the input features.
In this case, human readability is hindered by the length of the representation (e.g., a linear combination of 2 variables is more immediate than a combination of 5 variables) and its type (e.g., a linear combination is more comprehensible than a polynomial function).
The means to associate these representations with readability scores is still an open issue.

\subsubsection{Microlevel observations}

While the macrolevel knowledge complexity is quite easy to be assessed, the same does not hold at the microlevel, because of the variety of adopted rule notations.
Different SKE algorithms adopt different notations and each one comes with its proper conciseness and human-interpretability degree.
Unfortunately, it is not possible to convert the various formats into a canonical, comparable notation and for this reason, the problem of assigning a readability score to individual rules is a major open issue in the assessment of the extracted knowledge quality.

\subsection{User interaction}

Since a quantitative assessment of the microlevel readability cannot be straightforwardly defined, a preliminary readability score formulation could take into account user feedback.
User interactions may drive such a score according to a custom \emph{desideratum} or a set of \emph{desiderata}.
For instance, propositional and \mofn{} rules may be preferred over fuzzy and oblique rules, and at the same time propositional rules may be preferred over the \mofn{} notation.
This solution may be achieved by assigning an \emph{ad hoc} weight to each rule category.

The user interaction is also important when balancing different indicators concurring within the same scoring metric.
As an example, we can consider a scoring function for the knowledge provided by SKE techniques involving the 3 indicators described in Section 2 (predictive performance, readability, and completeness).
We recall that knowledge has good quality if it has high predictive performance, high readability, and high completeness.

An effective score may be formulated by multiplying 3 distinct, numerical metrics, one per indicator.
As a result, when performing a comparison, the knowledge having the highest score is the one having the best quality.
However, in this scenario, it is important to correctly weight the different indicators, and in particular to refine the fidelity/readability trade-off (i.e., how to penalise fidelity losses w.r.t.\ readability losses) based on the task at hand and according to the user experience.
This may be achieved by introducing a human parametrisation acting on the individual indicators, that will be finally merged into a single multiplicative scoring function.
It is relevant noticing that the completeness is not task-dependent, since it may be represented as a percentage.
Conversely, the other indicators may be subject to a large variability, depending on the task at hand.

\begin{figure*}[h]\centering
	\subfloat[Data set.]{
		\includegraphics[width=\threeinarow]{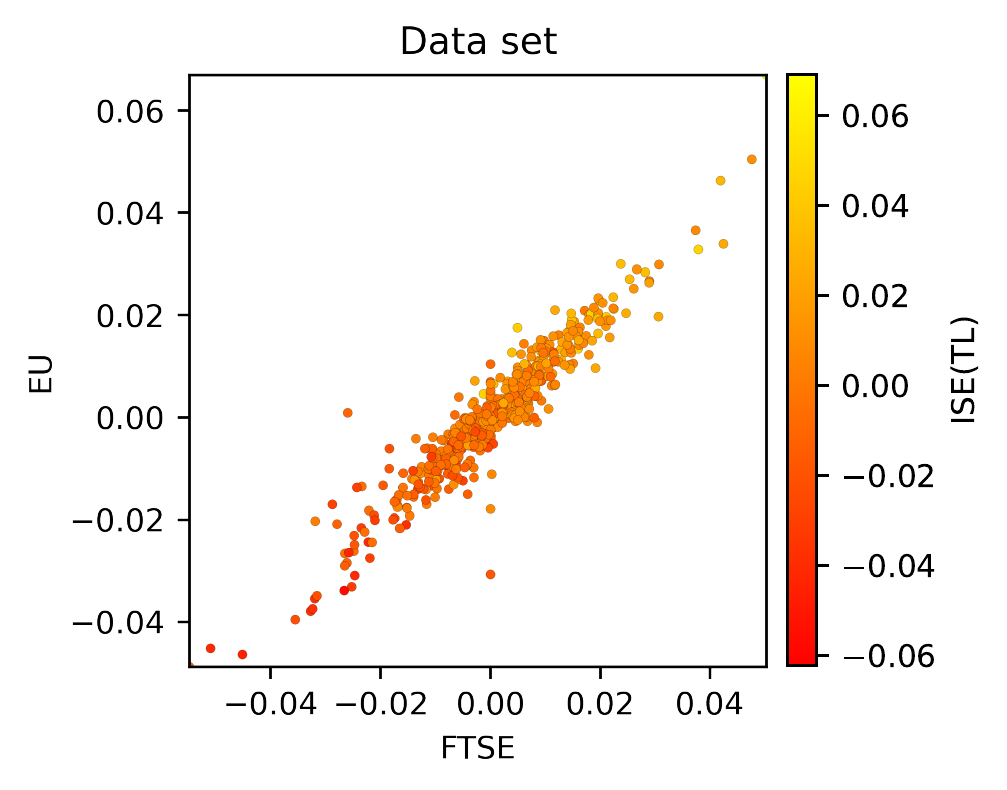}\label{fig:data}
	}
	\subfloat[Linear regressor (LR).]{
		\includegraphics[width=\threeinarow]{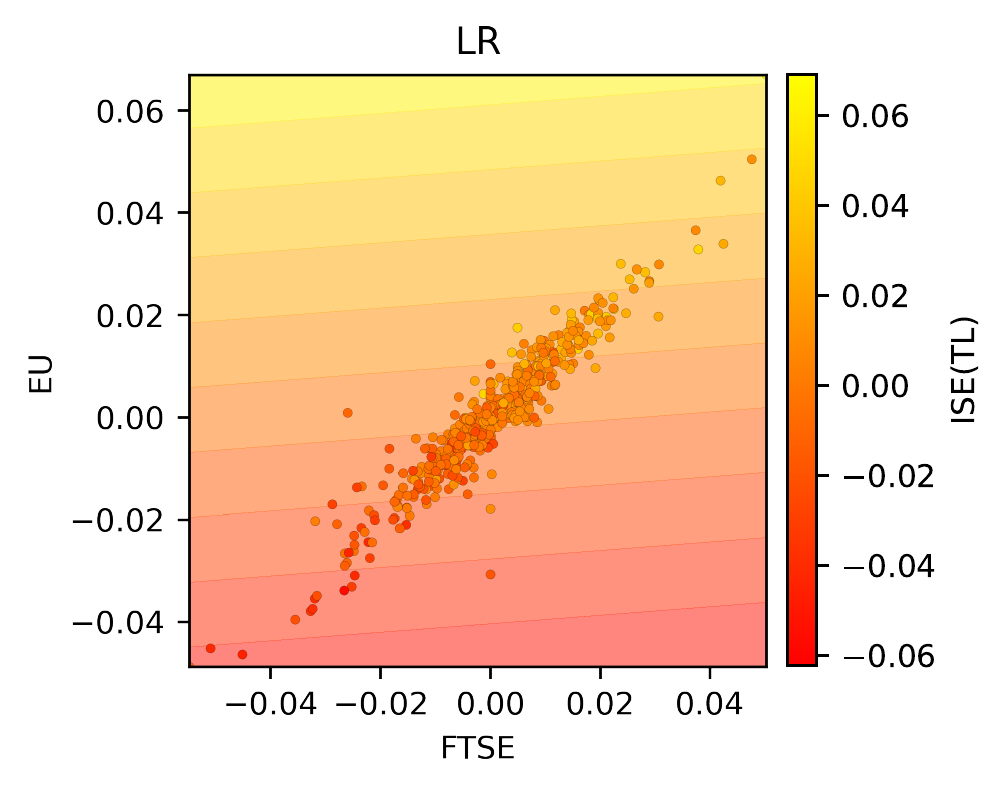}\label{fig:lr}
	}
	\subfloat[Random forest (RF).]{
		\includegraphics[width=\threeinarow]{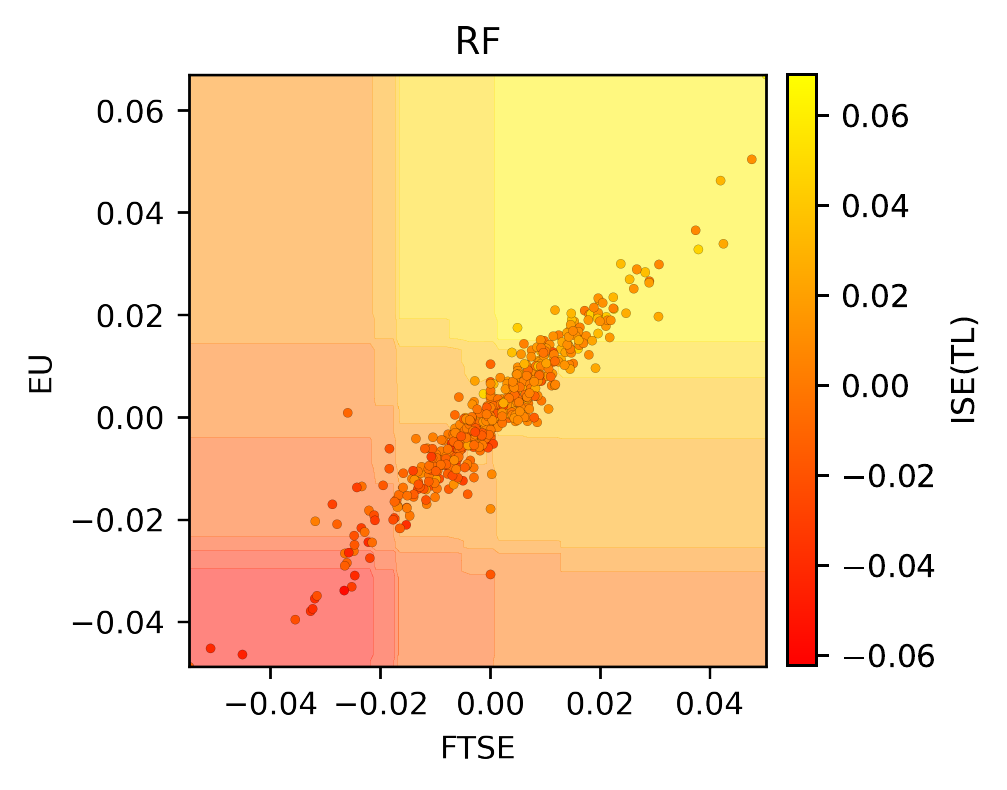}\label{fig:rf}
	}
	\caption{Data set used in the experiments and corresponding decision boundaries obtained by training a linear regressor and a random forest. Only the two most relevant input features are reported---i.e., the MSCI European index (EU) and the stock market return index of UK (FTSE).}\label{fig:dataPred}
\end{figure*}

\begin{figure*}[p]\centering
	\subfloat[\cart{} 6 leaves (LR).]{
		\includegraphics[width=\threeinarow]{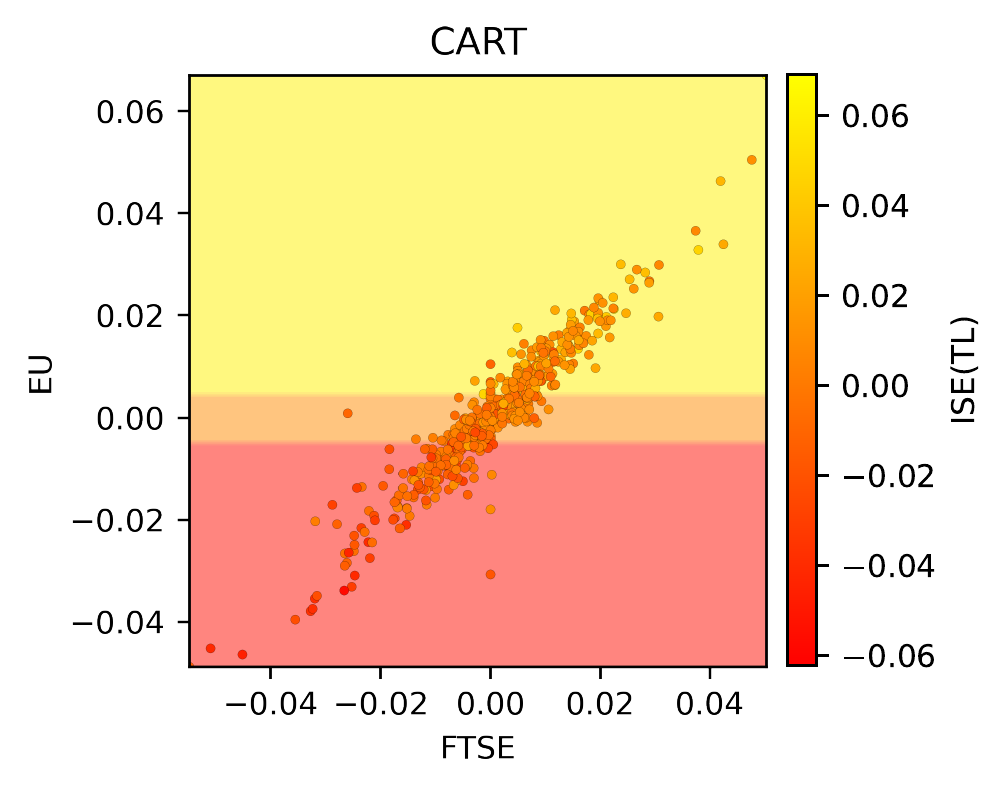}\label{fig:cart6}
	}
	\subfloat[\cart{} 10 leaves (LR).]{
		\includegraphics[width=\threeinarow]{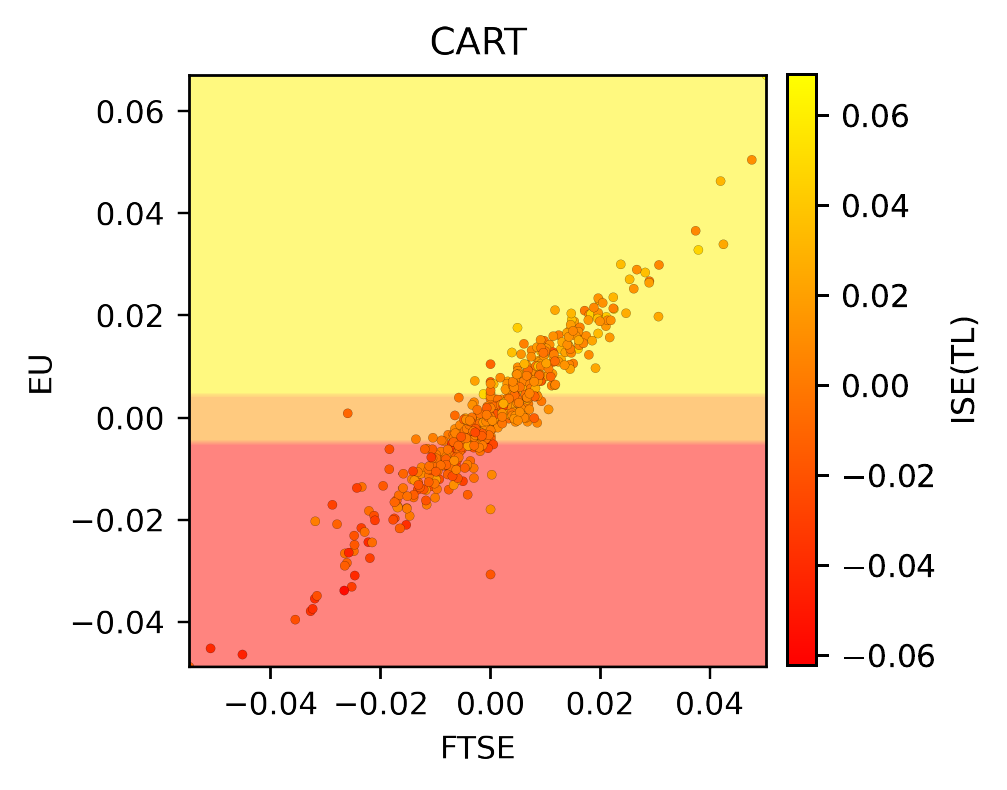}\label{fig:cart10}
	}
	\subfloat[\creepy{} constant output (LR).]{
		\includegraphics[width=\threeinarow]{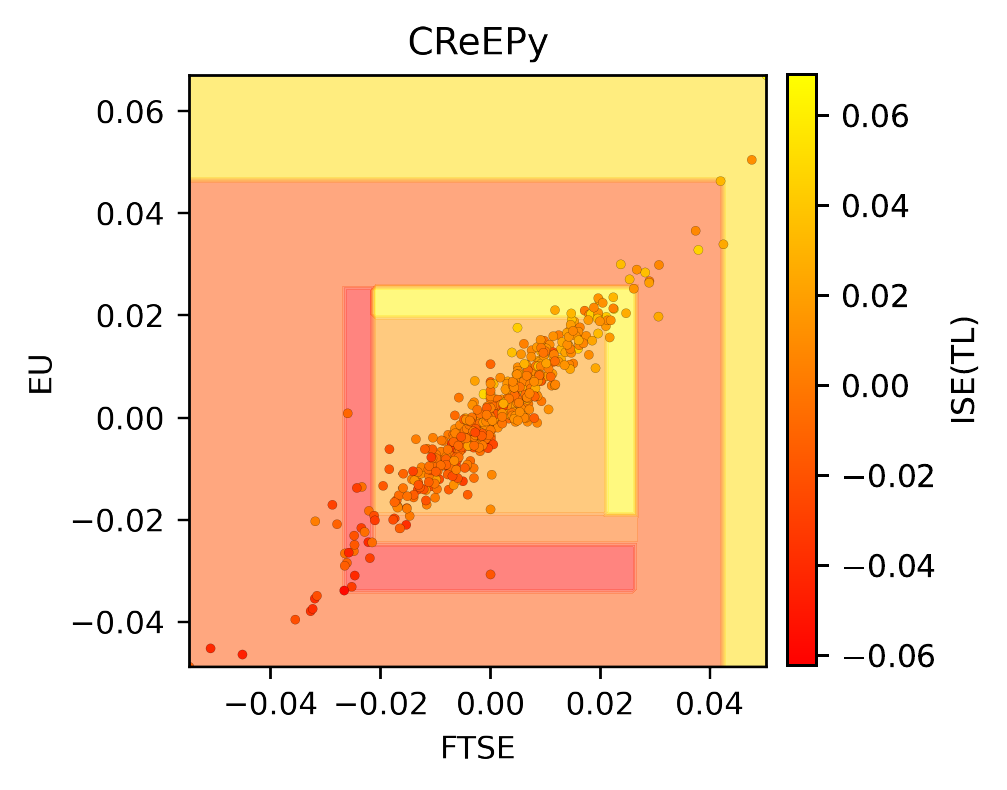}\label{fig:creepyK}
	}
	//
	\subfloat[\creepy{} (LR).]{
		\includegraphics[width=\threeinarow]{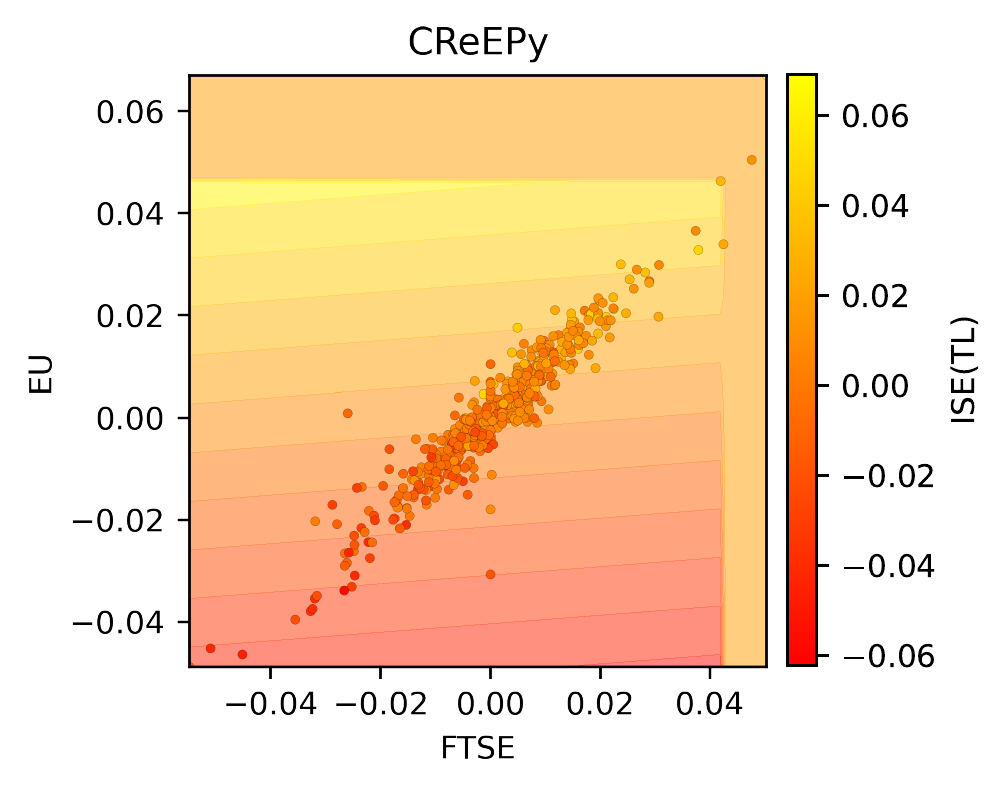}\label{fig:creepy}
	}
	\subfloat[\gridex{} (LR).]{
		\includegraphics[width=\threeinarow]{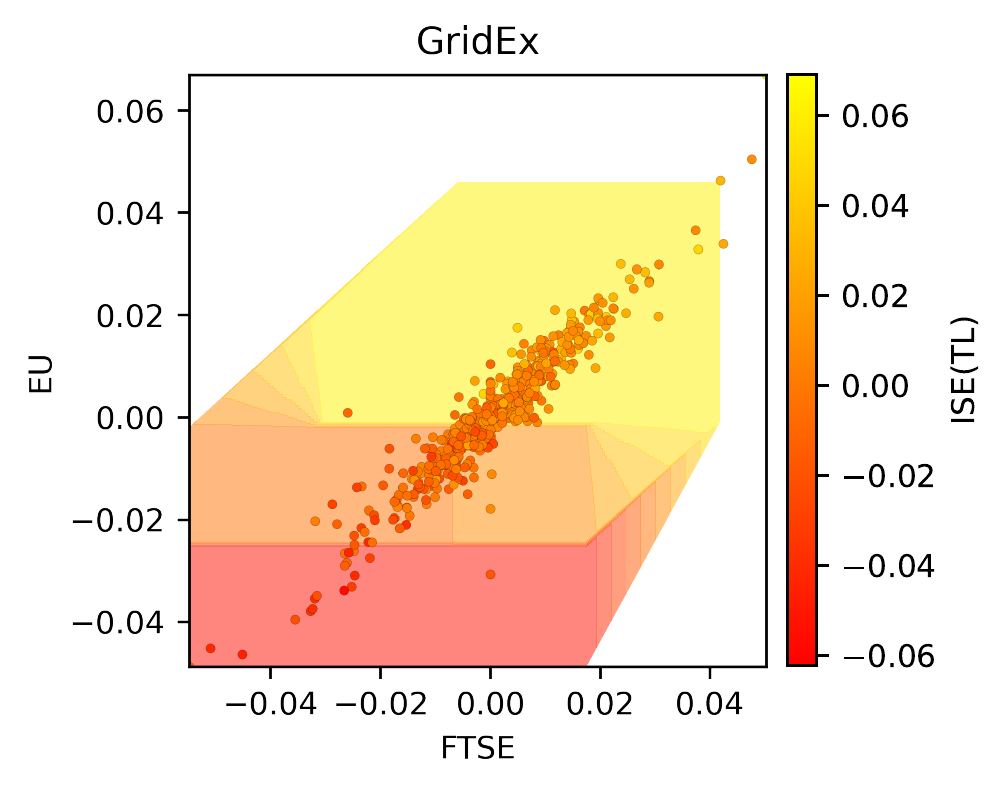}\label{fig:gridex}
	}
	\subfloat[\gridrex{} (LR).]{
		\includegraphics[width=\threeinarow]{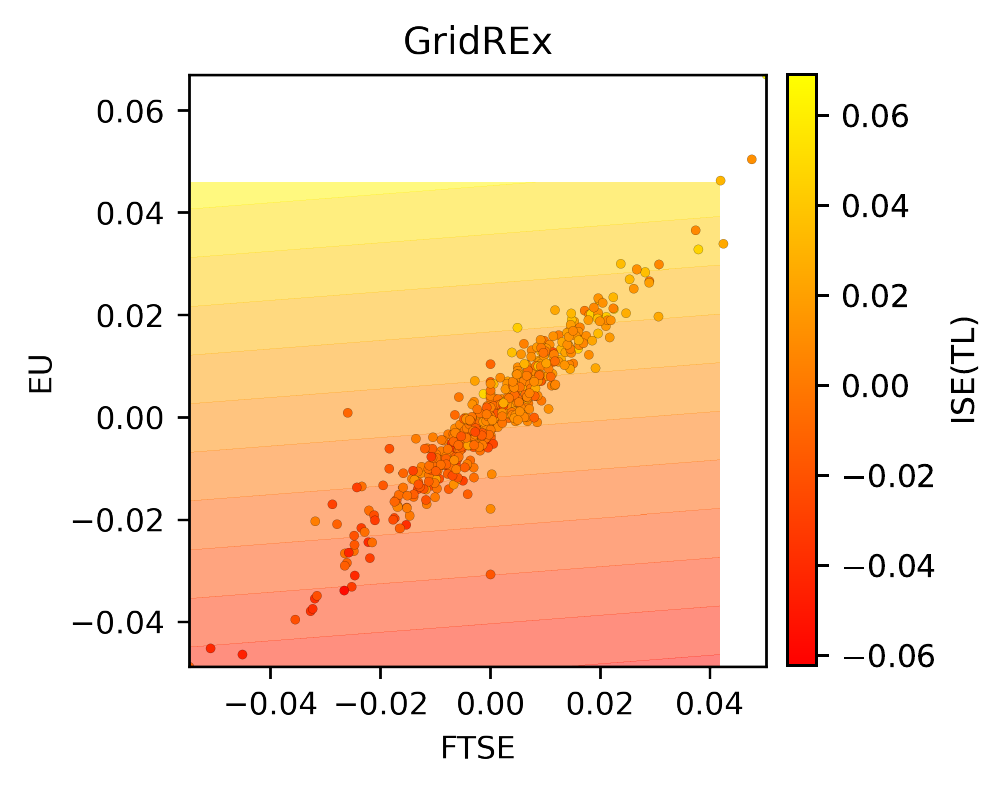}\label{fig:gridrex}
	}
	//
	\subfloat[\cart{} 6 leaves (RF).]{
		\includegraphics[width=\threeinarow]{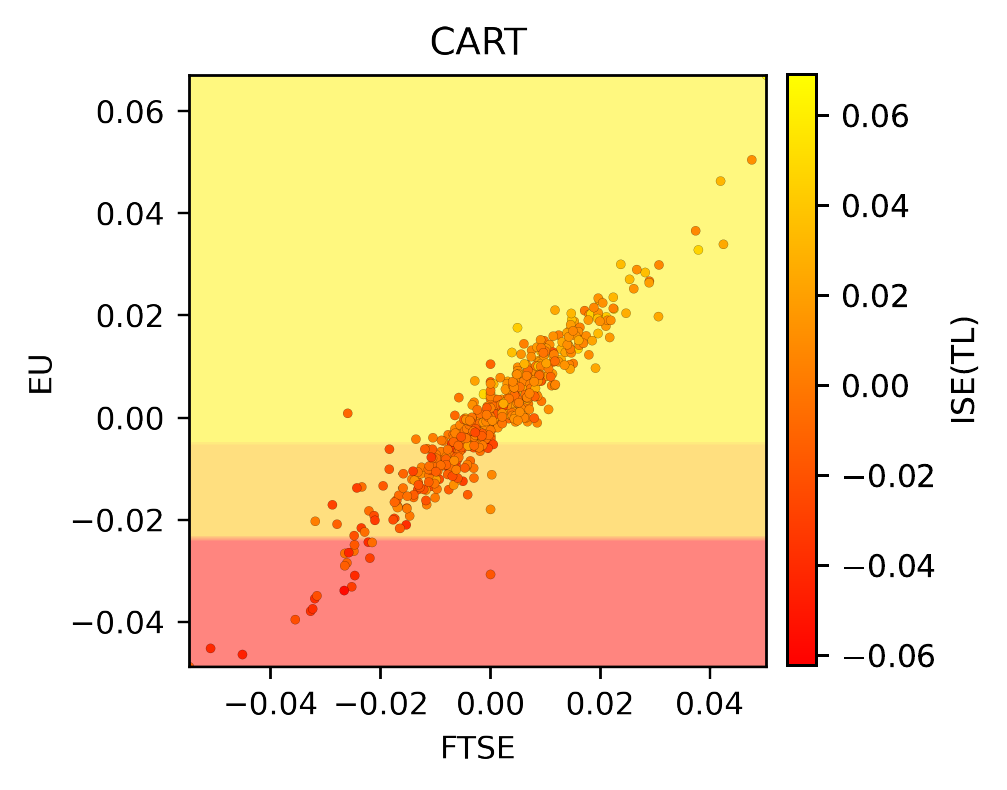}\label{fig:cart6RF}
	}
	\subfloat[\cart{} 10 leaves (RF).]{
		\includegraphics[width=\threeinarow]{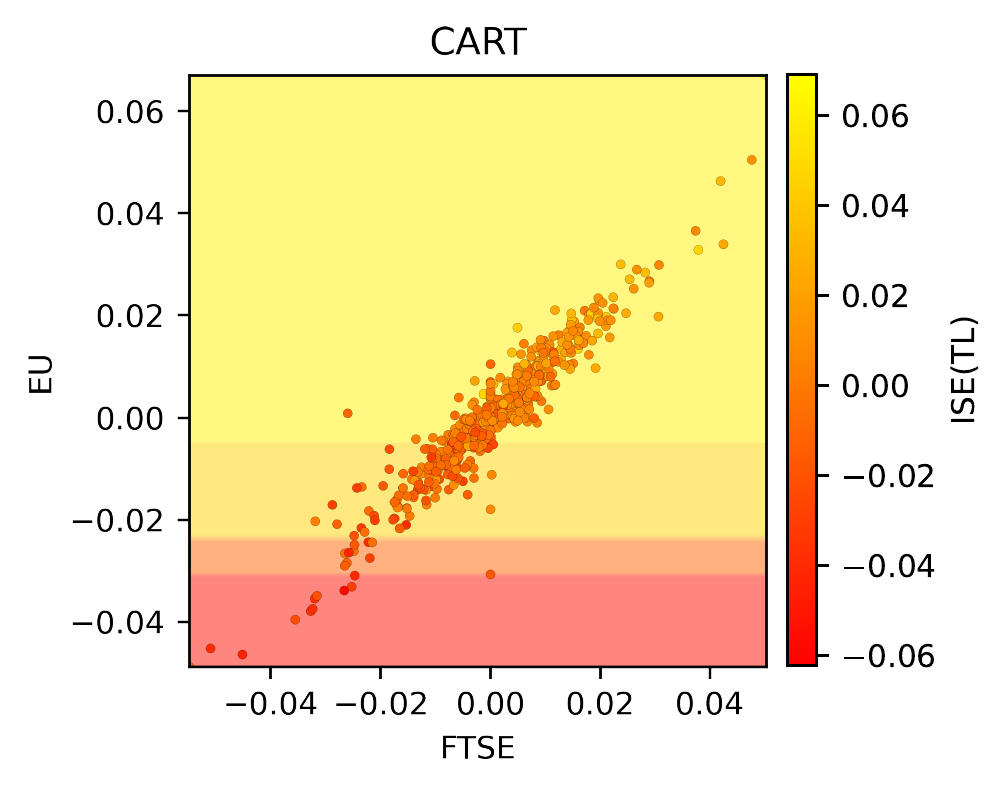}\label{fig:cart10RF}
	}
	\subfloat[\creepy{} constant output (RF).]{
		\includegraphics[width=\threeinarow]{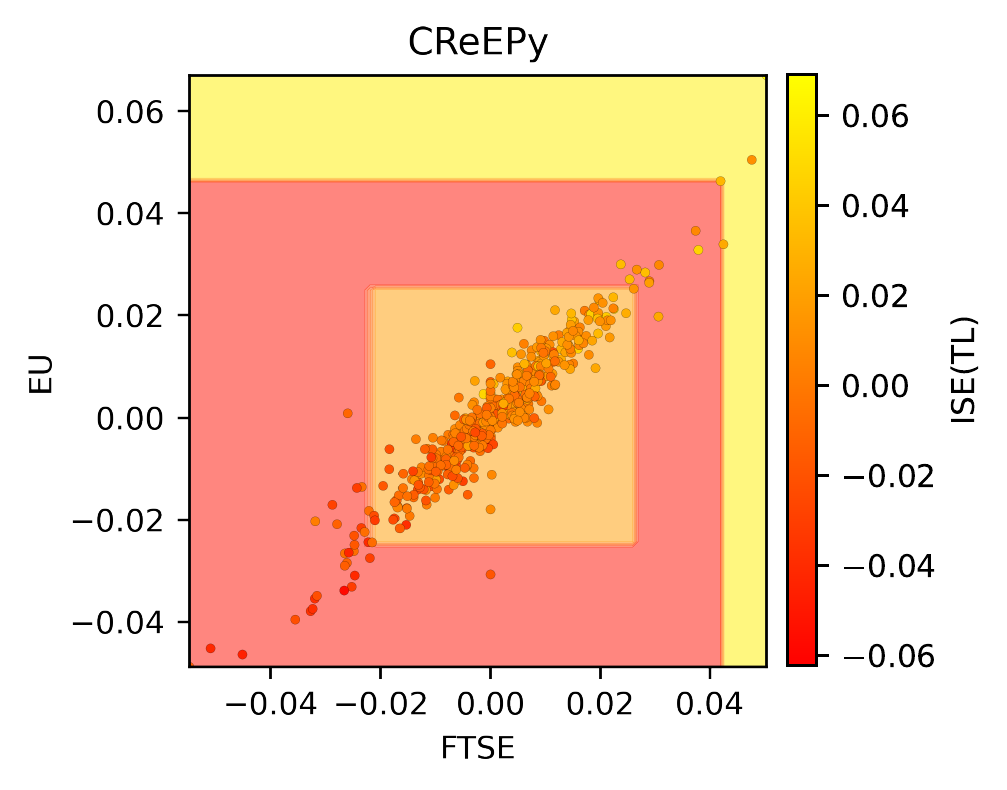}\label{fig:creepyKRF}
	}
	//
	\subfloat[\creepy{} (RF).]{
		\includegraphics[width=\threeinarow]{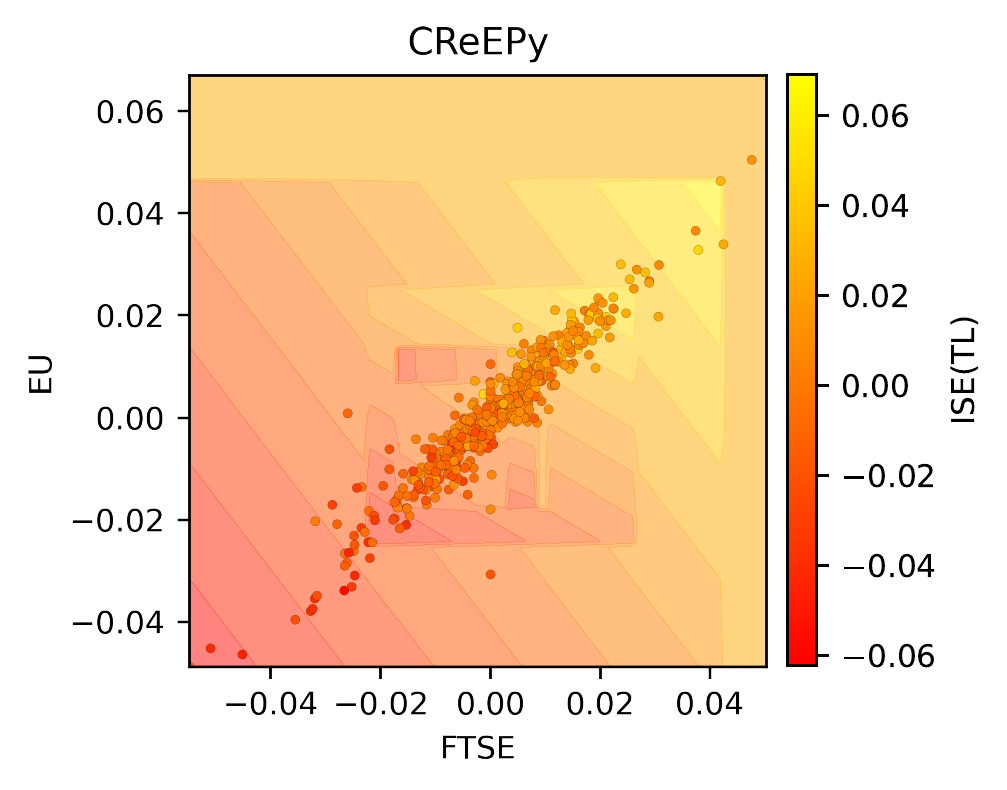}\label{fig:creepyRF}
	}
	\subfloat[\gridex{} (RF).]{
		\includegraphics[width=\threeinarow]{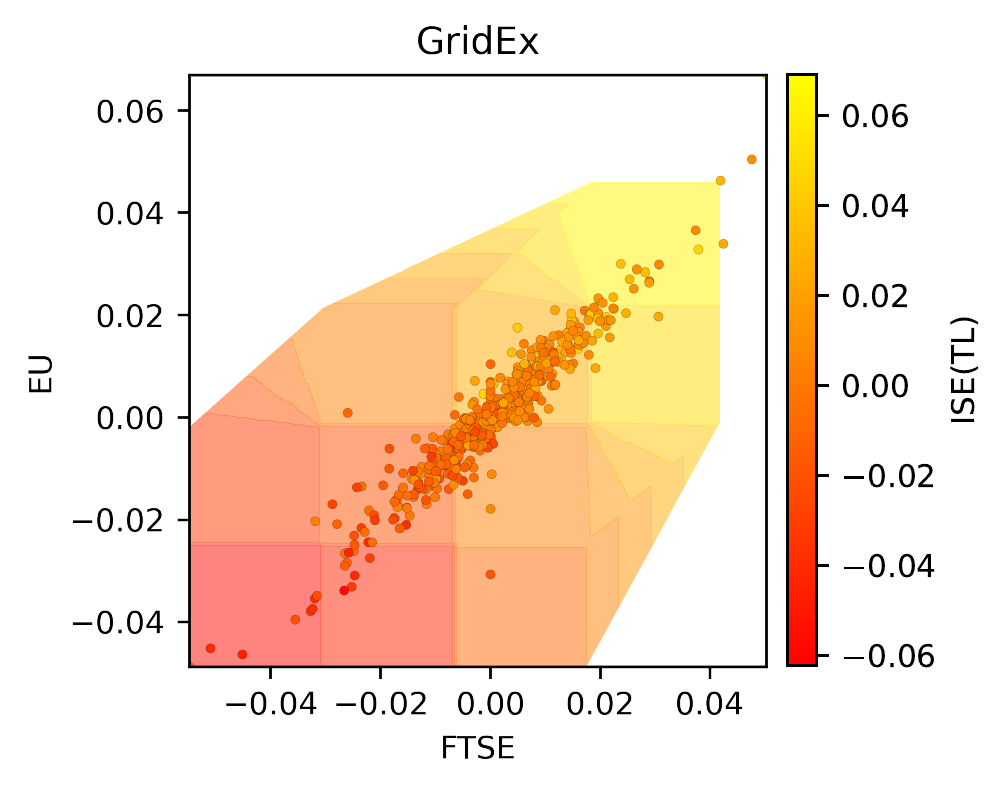}\label{fig:gridexRF}
	}
	\subfloat[\gridrex{} (RF).]{
		\includegraphics[width=\threeinarow]{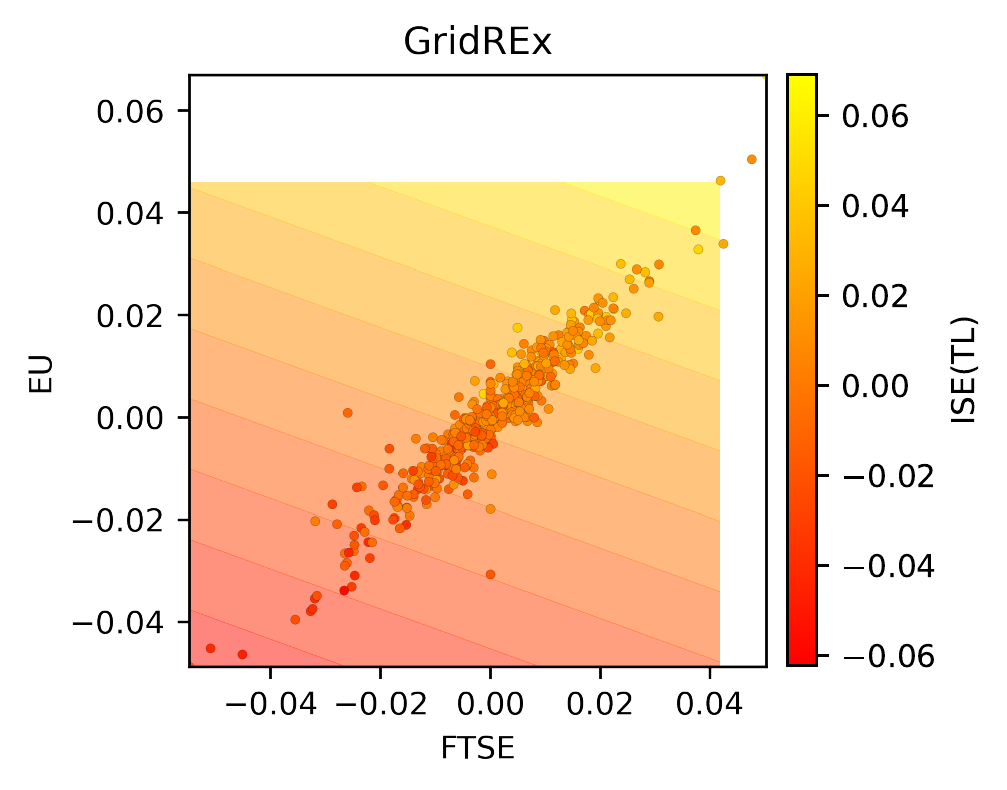}\label{fig:gridrexRF}
	}
	\caption{Decision boundaries for the data set shown in \Cref{fig:data} obtained via the application of different extractors to the LR and RF predictors (cf. \Cref{fig:lr,fig:rf}, respectively).}\label{fig:experiments}
\end{figure*}

\section{The Istanbul Stock Exchange Data Set Example}

\begin{table*}[t]\centering
	\caption{Quality assessments for the knowledge extracted by different SKE algorithms from a linear regressor and a random forest BB for the Istanbul Stock Exchange data set.}\label{tab:experiments}
	\begin{tabular}{cccccccc}
		\toprule
		\multicolumn{3}{c}{Predictor} &\multicolumn{5}{c}{Extractor} \\
		\cmidrule(lr){1-3}
		\cmidrule(lr){4-8}
		Type & MAE & R$^2$ & Type & Parameters & Rules & MAE data (BB) & R$^2$ data (BB) \\
		\midrule
		LR & 0.01 & 0.47 & \cart{} & Max leaves = 6 & 6 & 0.01 (0.00) & 0.42 (0.78) \\
		& & & & Max depth = Unbounded & & & \\
		\cmidrule(lr){4-8}
		& & & \cart{} & Max leaves = 10 & 10 & 0.01 (0.00) & 0.43 (0.85) \\
		& & & & Max depth = Unbounded & & & \\
		\cmidrule(lr){4-8}
		& & & \creepy{} & Max depth = 6 & 7 & 0.01 (0.01) & 0.10 (0.16) \\
		& & & & Threshold = 0.001 & & & \\
		& & & & Output = constant & & & \\
		\cmidrule(lr){4-8}
		& & & \creepy{} & Max depth = 6 & 2 & 0.01 (0.00) & 0.44 (0.93) \\
		& & & & Threshold = 0.001 & & & \\
		& & & & Output = linear combination & & & \\
		\cmidrule(lr){4-8}
		& & & \gridex{} & Max depth = 2 & 7 & 0.01 (0.01) & 0.30 (0.61) \\
		& & & & Threshold = 0.01 & & & \\
		& & & & Splits = 3 if feature importance > 0.8 else 1 & & & \\
		\cmidrule(lr){4-8}
		& & & \gridrex{} & Max depth = 2 & 1 & 0.01 (0.00) & 0.44 (1.00) \\
		& & & & Threshold = 0.01 & & & \\
		& & & & Splits = 2 & & & \\
		\midrule
		RF & 0.01 & 0.42 & \cart{} & Max leaves = 6 & 6 & 0.01 (0.00) & 0.38 (0.90) \\
		& & & & Max depth = Unbounded & & & \\
		\cmidrule(lr){4-8}
		& & & \cart{} & Max leaves = 10 & 8 & 0.01 (0.00) & 0.41 (0.90) \\
		& & & & Max depth = Unbounded & & & \\
		\cmidrule(lr){4-8}
		& & & \creepy{} & Max depth = 6 & 7 & 0.01 (0.01) & 0.03 (0.06) \\
		& & & & Threshold = 0.001 & & & \\
		& & & & Output = constant & & & \\
		\cmidrule(lr){4-8}
		& & & \creepy{} & Max depth = 6 & 7 & 0.01 (0.01) & 0.40 (0.90) \\
		& & & & Threshold = 0.001 & & & \\
		& & & & Output = linear combination & & & \\
		\cmidrule(lr){4-8}
		& & & \gridex{} & Max depth = 2 & 10 & 0.01 (0.01) & 0.28 (0.62) \\
		& & & & Threshold = 0.01 & & & \\
		& & & & Splits = 3 if feature importance > 0.8 else 1 & & & \\
		\cmidrule(lr){4-8}
		& & & \gridrex{} & Max depth = 2 & 1 & 0.01 (0.00) & 0.41 (0.95) \\
		& & & & Threshold = 0.1 & & & \\
		& & & & Splits = 2 & & & \\
		\bottomrule
	\end{tabular}
\end{table*}

To show in practice the differences amongst the output of different knowledge extractors, we applied the \cart{}, \creepy{}, \gridex{} and \gridrex{} extractors to the Istanbul Stock Exchange data set\footnote{\vurl{https://archive.ics.uci.edu/ml/datasets/ISTANBUL+STOCK+EXCHANGE}}, reported in \Cref{fig:data}.
The data set is composed of 8 continuous features (7 as input + 1 as output).
We considered all of them during the training phase of black boxes and extractors, however, for the sake of readability,  we reported in \Cref{fig:dataPred,fig:experiments} only the most relevant two---i.e., the MSCI European index (EU) and the stock market return index of UK (FTSE).

The selected extractors have been applied to a linear regressor (LR) and a random forest (RF) BB.
Mean absolute error (MAE) and R$^2$ score of the BB predictions are reported in \Cref{tab:experiments}.
Corresponding decision boundaries are visually represented in \Cref{fig:lr,fig:rf}.
Predictions obtained through the extracted knowledge have been evaluated with the same scores adopted for the underlying BB outputs and are also reported in \Cref{tab:experiments}.
The corresponding decision boundaries are shown in Figures \ref{fig:cart6} to \ref{fig:gridrexRF}.

Considerations about the macrolevel knowledge readability can be drawn by analysing the plots in \Cref{fig:experiments}.
First of all, it is noticeable that \cart{}, \gridex{} and \creepy{} with constant outputs introduce an undesired discretisation in the output predictions, thus implying a worsening in the overall predictive performance.
Conversely, \gridrex{} and \creepy{} with linear outputs can provide linear combinations of the input variables as output predictions, resulting in knowledge having fewer rules with smaller predictive errors.

Observing the same plots it is noticeable that \creepy{} outputs a hierarchical knowledge, thus mappable in an ordered list of rules.
The same holds for \cart{}'s knowledge -- if simplified -- since it is based on a decision tree.
Differently, \gridex{} and \gridrex{} provide rules associated with hypercubic and non-overlapping regions, so they are not ordered and each one contains the same amount of variable occurrences, predicates (i.e., interval inclusion), and constant values (i.e., 2 per variable).

Macrolevel readability comparisons can be easily performed by observing the `Rules' column in \Cref{tab:experiments}.
\gridrex{} is the best choice since it outputs a single rule, followed by the \creepy{}. 
Their superiority is confirmed also from the predictive performance standpoint, since they have the highest R$^2$ score.
\gridrex{} doesn't meet the completeness requirement, i.e., results are non-exhaustive in the input space coverage since a portion is considered negligible due to the absence of training instances (so there are test instances that cannot be predicted). This should penalise the final metric score.

To perform microlevel comparisons it is necessary to observe the output knowledge rules provided by the various extractors.
We exemplify the difference between constant and linear outputs by considering the knowledge extracted via \cart{} and \creepy{} from the linear regressor.
\cart{} trained with a maximum of 6 leaves provides the following rules:
\begin{lstlisting}
ise(BOVESPA, DAX, EM, EU, FTSE, NIKKEI, SP, -0.02) :-
   EU =< -0.00, EU =< -0.02.
ise(BOVESPA, DAX, EM, EU, FTSE, NIKKEI, SP, 0.00) :-
   EU =< -0.00, EU > -0.02.
ise(BOVESPA, DAX, EM, EU, FTSE, NIKKEI, SP, 0.00) :-
   EU > -0.00, EU =< 0.01, EM =< 0.00.
ise(BOVESPA, DAX, EM, EU, FTSE, NIKKEI, SP, 0.00) :-
   EU > -0.00, EU =< 0.01, EM > 0.00.
ise(BOVESPA, DAX, EM, EU, FTSE, NIKKEI, SP, 0.01) :-
   EU > -0.00, EU > 0.01, EM =< 0.01.
ise(BOVESPA, DAX, EM, EU, FTSE, NIKKEI, SP, 0.02) :-
   EU > -0.00, EU > 0.01, EM > 0.01.
\end{lstlisting}
\cart{} rules may be simplified in the following without modifying their semantics:
\begin{lstlisting}
ise(BOVESPA, DAX, EM, EU, FTSE, NIKKEI, SP, -0.02) :-
   EU =< -0.02.
ise(BOVESPA, DAX, EM, EU, FTSE, NIKKEI, SP, 0.00) :-
   EU =< -0.00.
ise(BOVESPA, DAX, EM, EU, FTSE, NIKKEI, SP, 0.00) :-
   EU =< 0.01, EM =< 0.00.
ise(BOVESPA, DAX, EM, EU, FTSE, NIKKEI, SP, 0.00) :-
   EU =< 0.01.
ise(BOVESPA, DAX, EM, EU, FTSE, NIKKEI, SP, 0.01) :-
   EM =< 0.01.
ise(BOVESPA, DAX, EM, EU, FTSE, NIKKEI, SP, 0.02).
\end{lstlisting}
Rules extracted with \creepy{} (linear output) for the Istanbul Stock Exchange data set are the following:
\begin{lstlisting}
ise(BOVESPA, DAX, EM, EU, FTSE, NIKKEI, SP, ISE) :-
   FTSE in [-0.05, 0.04], EU in [-0.04, 0.04], ISE is 
   -0.02 BOVESPA - 0.05 DAX - 0.05 EM + 0.63 NIKKEI + 0.47 SP.
ise(BOVESPA, DAX, EM, EU, FTSE, NIKKEI, SP, ISE) :- ISE is 0.01.
\end{lstlisting}

On one hand, the two versions of the \cart{}'s knowledge are equivalent in terms of predictive accuracy and input space coverage, i.e., they provide the same predictions.
However, the first has redundant literals, whereas the second has implicit constraints.
This means that, when measuring the knowledge microlevel readability, the first should be penalised for the presence of literals \emph{in single clauses} that are trivially true.
For instance, the conjunction \verb|EU > -0.00, EU > 0.01, EM =< 0.01| hinders readability if compared to the equivalent \verb|EU > 0.01, EM =< 0.01|.
\emph{Vice versa}, the simplified version of the same rule represented as \verb|EM =< 0.01| implies implicitly acknowledging as false the constraint \verb|EU =< 0.01|, requiring humans to check all the previous rules.
An ideal microlevel readability metric should be able to assign a good score to \verb|EU > 0.01, EM =< 0.01| as well as penalise the redundant and implicit alternatives.

On the other hand, by comparing \cart{}'s and \creepy{}'s extracted knowledge it is clear that they exhibit
\begin{inlinelist}
	\item exactly the same input space coverage;
	\item similar predictive performance (\creepy{} is slightly better in the fidelity w.r.t.\ the BB predictions);
	\item different macrolevel readability, since \creepy{} extracts only 2 rules instead of 6, as \cart{};
	\item different microlevel readability, since \creepy{}'s predictions may be constants as well as linear combinations of up to 7 variables.
\end{inlinelist} 
An ideal microlevel readability metric should be able to assign a score taking into account these differences.
The lack of a quantitative metric to evaluate the microlevel readability -- and thus the overall knowledge readability and, in turn, quality -- makes it impracticable to automatically decide whether longer rule lists with simple outputs are preferable or not over shorter lists with more complex outputs.

\section{Conclusions \& discussion}

In this paper we discuss the major issues in defining possible metrics to evaluate in a quantitative way the \emph{quality} of the symbolic knowledge extracted via SKE techniques out of a BB model as well as the need to define such metrics.
We show that quality evaluation should consider different indicators, namely: \textit{i)} predictive performance, \textit{ii)} input space coverage and \textit{iii)} knowledge readability.
While for the first two the definition of a metric and the comparison of a score is quite trivial, this does not happen for the last one.
As for the latter, readability can be observed at a macrolevel -- as the shape and complexity of the knowledge as a whole -- and at a microlevel---as shape and complexity of each knowledge rule.
We showed that at a macrolevel the knowledge may be easily compared since it is quite easy to convert one format into another.
Conversely, we pointed out the most tricky characteristics of microlevel considerations, to favour further studies and discussion on the topic.

Future works will be devoted to the definition of such a suitable metric to assess the quality of knowledge extracted via SKE in a fair, quantitative, and measurable way.
In particular, we plan to formulate a sound scoring function for microlevel comparisons and to study how the user interaction may help the customisation and integration of the 3 aforementioned indicators in a unique and effective metric assessing the knowledge quality.

Moreover, adherence to the trustworthy AI requirements should be enlarged to evaluate also other requirements and, in particular, our findings might be subject to change due to the institutional debate about the AI Act Proposal. Further research is needed to consolidate the interpretation of the Act in light of its future changes and to define metrics that can be used in the course of the standardisation process while being respectful of the AI Act and its Annexes.

Another open question left to further research is to take into account that the definition of explainability metrics may lead to consider concurring features that are compliant with the legislative goals of the AI Act and, eventually, the thresholds that meet the Act’s requirements and expectations.
Each feature needs to be balanced via parameters that can be set taking into consideration both the scenario at hand and its regulation. This is the first step for converging to trustworthy automated machine learning techniques that we believe to be promising in the near future.

\begin{acks}
This work has been partially supported by the EU ICT-48 2020 project TAILOR (No. 952215) and by the European Union’s Horizon 2020 research and innovation programme under G.A. no. 101017142 (StairwAI project).
\end{acks}

\bibliographystyle{ACM-Reference-Format}
\bibliography{icaif-2022-scoring}

\end{document}